\documentclass[letterpaper, 10 pt, conference]{ieeeconf}  

\IEEEoverridecommandlockouts                              

\usepackage{graphics} 
\usepackage{epsfig} 
\usepackage{times} 
\usepackage{amsmath} 
\usepackage{xcolor}

\usepackage{mathrsfs}
\usepackage{amssymb}
\usepackage{cite}
\usepackage{dsfont}
\usepackage{xcolor}
\usepackage{mathtools}
\usepackage{mathrsfs}
\usepackage{subfig}
\usepackage{balance}
\usepackage{float}
\usepackage{flushend}
\newtheorem{remark}{Remark}

\newcommand{\rom}[1]{%

\textup{\uppercase\expandafter{\romannumeral#1}}%
}
\title{\LARGE 
Exact Imposition of Safety Boundary Conditions in Neural Reachable Tubes
\vspace{-0.5em}
}

\author{Aditya Singh$^*$$^{1,}$$^{2}$, Zeyuan Feng$^*$$^{1,3}$, and Somil Bansal$^{1,}$$^{3}$
\thanks{$^*$ The first two authors contributed equally to this work.
}%
\thanks{$^{1}$Authors are with the ECE Department at University of Southern California, LA, USA:
        {\{zeyuanfe, somilban\}@usc.edu.}}%
\thanks{$^{2}$Author is with the EE Department at Indian Institute of Technology Patna, India:
        {\{aditya\_2001ee04@iitp.ac.in\}}.}%
\thanks{$^{3}$Authors are with the Department of Aeronautics and Astronautics at Stanford University, USA {:\{zeyuanf, somil\}@stanford.edu}.}%
\thanks{This research is supported in part by the DARPA Assured Neuro Symbolic Learning and Reasoning (ANSR) program and by the NSF CAREER program (2240163).}%
}

\begin{document}
\maketitle
\thispagestyle{empty}
\pagestyle{empty}

\begin{abstract}

Hamilton-Jacobi (HJ) reachability analysis is a widely adopted verification tool to provide safety and performance guarantees for autonomous systems.
However, it involves solving a partial differential equation (PDE) to compute a safety value function, whose computational and memory complexity scales exponentially with the state dimension, making its direct application to large-scale systems intractable. 
To overcome these challenges, DeepReach, a recently proposed learning-based approach, approximates high-dimensional reachable tubes using neural networks (NNs). 
While shown to be effective, the accuracy of the learned solution decreases with system complexity. 
One of the reasons for this degradation is a soft imposition of safety constraints during the learning process, which corresponds to the boundary conditions of the PDE, resulting in inaccurate value functions.
In this work, we propose ExactBC, a variant of DeepReach that imposes safety constraints \textit{exactly} during the learning process by restructuring the overall value function as a weighted sum of the boundary condition and the NN output. 
Moreover, the proposed variant no longer needs a boundary loss term during the training process, thus eliminating the need to balance different loss terms. 
We demonstrate the efficacy of the proposed approach in significantly improving the accuracy of the learned value function for four challenging reachability tasks: a rimless wheel system with state resets, collision avoidance in a cluttered environment, autonomous rocket landing, and multi-aircraft collision avoidance.
\end{abstract}

\section{Introduction}
With the recent surge in real-world applications of autonomous systems, ensuring the safety of these systems is becoming increasingly important. One effective approach to formalize the safety of autonomous systems is via Hamilton-Jacobi (HJ) Reachability analysis \cite{8263977}, wherein a Backward Reachable Tube (BRT) of the system is computed. 
The BRT is the set of all initial states of the system from which it will eventually enter a failure set despite its best control effort. 
Therefore, the BRT represents the unsafe states for the system and should be avoided. Along with the BRT, reachability analysis also yields a safe controller for the system that keeps it outside the BRT and inside the safe set.
Conversely, the reachability analysis and BRT can also be used to ensure liveness, given a target set. 

The BRT computation in HJ reachability ultimately amounts to computing a value function by solving a partial differential equation (PDE) called Hamilton-Jacobi-Bellman (HJB) PDE.
Once computed, the sub-zero level set of the value function yields the BRT. 
Traditionally, this PDE is solved numerically over a state-space grid \cite{mitchell2005time,lygeros2004reachability,fisac2015reach,liao2024improved}. However, the computational complexity of grid-based methods scales exponentially with the number of state dimensions, limiting their direct application to systems up to 5 dimensions.

Several recent advancements have been made to mitigate this ``curse of dimensionality.'' Some approaches impose assumptions on system dynamics, such as linearity \cite{darbon2016algorithms,chow2017algorithm,lee2020hopf} or monotonicity \cite{coogan2015efficient}, or decompose the system dynamics to reduce a high-dimensional reachability problem to several smaller sub problems \cite{chen2015exact,10365682}. However, these methods do not offer a general framework for handling high-dimensional system dynamics within HJ reachability analysis. To overcome these challenges, learning-based methods have been proposed to synthesize neural approximations of the reachable tubes and the value function \cite{9561949, darbon2020overcoming, niarchos2006neural,djeridane2006neural,rubiesroyo2017recursive,jiang2016using}. 
Among these, a promising class of methods leverages deep neural networks (DNNs) to approximate the safety value function, such as DeepReach \cite{9561949}. 
Inspired by recent advances in physics-informed machine learning \cite{raissi2019physics, saad2022guiding,li2020fourier,li2021physics}, DeepReach represents the safety value function as a NN whose weights are optimized using a combination of two loss terms: (a) the residual errors arising from enforcing the boundary conditions of the HJB PDE, which corresponds to the specification of the failure set (also referred to as the \textit{safety constraint} from here on) and (b) the residual errors associated with the HJB PDE within the domain’s interior. 
Hence, the accuracy of the learned solution directly depends on the ability to co-optimize the two loss terms and the relative weight between them. 
Additionally, the imposition of safety conditions as a soft constraint makes DeepReach particularly prone to learning inaccurate value functions, as the solution of any PDE depends heavily on its boundary conditions. 
This drawback is more acute for complex failure sets (e.g., cluttered obstacles for a mobile robot), where DeepReach may learn completely non-physical solutions.

In this work, we propose ExactBC (an acronym for \underline{Exact} \underline{B}oundary \underline{C}onditions), a new variant of DeepReach that exactly satisfies the safety boundary conditions. Inspired by \cite{ sukumar2022exact, lagaris1998artificial}, ExactBC represents the value function as a weighted combination of the boundary value function and the NN output, where the network output is weighted by a ``distance function'' to the boundary. 
This distance function is zero at the boundary and increases as the system state is farther from the boundary. 
Consequently, we show that the learned value function always satisfies the boundary conditions. 
Corresponding to this novel formulation, we derive a new loss function to train the NN.
Since the network needs to optimize only the residual PDE errors under ExactBC, it eliminates the scope of inaccuracies that could stem from the incorrect weighting of two loss terms in DeepReach. 
We show that ExactBC significantly enhances the accuracy of the learned safety value function while simplifying the learning process.
To summarize, our overall contributions are:
\begin{enumerate}
    \item We propose ExactBC, a new method to exactly impose safety boundary conditions in deep learning-based reachability methods.
    \item We illustrate the benefits of ExactBC on four challenging reachability problems of varying dimensionality and complexity of safety constraints.
    We demonstrate that ExactBC significantly increases the accuracy of the learned value functions and eliminates the need to optimize any hyperparameters in the loss function.
\end{enumerate}
\section{Problem Setup} \label{Problem Setup}
We consider a nonlinear, time-invariant dynamical system
$\dot{x}=f(x,u)$, where $x \in \mathbb{R}^n$ is the system state and $u \in \mathcal{U} \subset \mathbb{R}^{n_u}$ is the control input.
%
%
We use $\xi^{u}_{x,t}(\tau)$ to denote the system state at time $\tau$, starting from an initial state $x$ at time $t$ on applying the control function $u(\cdot)$ over the horizon $[t, \tau]$.
%

We are given a \textit{failure set} $\mathcal{L}$ that represents the undesirable states for the system (e.g., obstacles for a mobile robot). 
Given $\mathcal{L}$, our \textit{goal} is to compute the BRT of the system, $\mathcal{B}(t)$, that represents all initial states that are doomed to fail:
%
\begin{equation}
\label{eq:avoid_BRT}
    \mathcal{B}(t)=\left\{x: \forall u(\cdot) \in \mathcal{U}_{[t, T]}, \exists \tau \in[t, T], \xi_{x, t}^{u}(\tau) \in \mathcal{L} \right\},
\end{equation}
where $\mathcal{U}_{[t, T]}$ denote all control functions over the time horizon $[t, T]$.
Along with the BRT, we want to compute a safety controller $\pi^*(x)$ for the system that keeps it outside the BRT. 
Conversely, if $\mathcal{L}$ represents a \textit{target set} (e.g., a goal region), then BRT represents all initial states from which liveness can be guaranteed:
%
\begin{equation}
\label{eq:reach_BRT}
    \mathcal{G}(t)=\left\{x: \exists u(\cdot) \in \mathcal{U}_{[t, T]}, \exists \tau \in[t, T], \xi_{x, t}^{u}(\tau) \in \mathcal{L} \right\}.
\end{equation}
In this case, along with $\mathcal{G}(t)$, we want to compute a liveness-preserving controller for the system that keeps the system inside $\mathcal{G}(t)$.
%
%
%
In this work, we will propose a learning-based method to compute the BRTs for both cases, $\mathcal{G}$ and $\mathcal{B}$. 

\section{Background: Hamilton-Jacobi Reachability}
Hamilton-Jacobi (HJ) Reachability analysis \cite{mitchell2005time, lygeros2004reachability} formulates the computation of BRTs as an optimal control problem. 
%
%
%
The first step is to define a Lipschitz-continuous target function $l: \mathbb{R}^n \rightarrow \mathbb{R}$ whose sub-zero level set is the failure set: $\mathcal{L}=\{x: l(x) \leq 0\}$.
A common choice for $l$ in robotics is the signed distance function to $\mathcal{L}$.
%
Given $l$, the BRT is obtained by computing the minimum distance to $\mathcal{L}$ throughout the system trajectory under the optimal control:
\vspace{-0.5em}
\begin{equation}
  V(x,t)  = \sup_{u(\cdot)} ~ \min _{\tau \in[t, T]} l\left(\xi_{x, t}^{u}(\tau)\right).
  \vspace{-0.5em}
\end{equation}
Intuitively, the optimal control tries to avoid and maximize the distance from the failure region. 
The sign of the value function $V(\cdot)$ indicates whether the system entered the failure set under this optimal control. Thus, the BRT is given by the sub-zero level set of the value function:
%
%
\vspace{-0.5em}
\begin{equation} \label{eq:BRT_from_value}
\mathcal{B}(t)=\left\{x: V(x,t)  \leq 0 \right\}.
\end{equation}

The value function can be computed via solving the following Hamilton-Jacobi-Bellman Variational Inequality (HJB-VI), which yields an identical solution as the HJB PDE while empirically providing a stronger learning signal for the NN:
\begin{equation}
\begin{gathered}
\label{eq: HJI-VI}
     \min \{D_{t}V(x,t) + H(x,t), l(x) - V(x,t) \} = 0, \\
     V(x,T) = l(x),\\
    H(x,t) = \max_{u \in \mathcal{U}} \langle \nabla V(x,t) \; , \; f(x,u) \rangle, 
\end{gathered}
\end{equation}
where $\nabla$ and $D_t$ denote the spatial and time derivatives of the value function.
$H(x,t)$ is the Hamiltonian of the system that encodes the role of control and how it affects the value function.
The boundary condition for the HJB-VI is given by $l(x)$, which defines the safety constraint for the system. 
For a detailed explanation and derivation, we refer interested readers to \cite{lygeros2004reachability, bokanowski2010reachability, 8263977}.

Once the value function is computed, the BRT is given by \eqref{eq:BRT_from_value}. 
Along with the BRT, the value function also provides an optimal safe policy for the system:
\begin{equation} \label{eqn:opt_ctrl}
    \pi^*(x,t)=\arg \max _u \langle\nabla V(x, t), f(x, u)\rangle .
\end{equation}
Intuitively, the safety controller steers the system towards higher values (i.e., away from the failure set) at any state $x$.

Conversely, when $\mathcal{L}$ denotes a target set, the same framework can be used to compute the BRT, except that the control minimizes the Hamiltonian in \eqref{eq: HJI-VI} and \eqref{eqn:opt_ctrl}.
%

\subsection{DeepReach}\label{DeepReach}
Traditionally, numerical methods are employed to solve the HJB-VI over a grid representation of the state space \cite{mitchell2004toolbox, pythonhjtoolbox}, wherein the time and spatial derivatives are approximated numerically over the grid to obtain a solution to the HJB PDE.
While the grid-based methods offer accurate solutions for low-dimensional problems, they suffer from the curse of dimensionality. Consequently, learning-based methods, such as DeepReach \cite{9561949}, have been developed to solve HJB-VI for high-dimensional cases. 
DeepReach leverages a DNN, parameterized by $\theta$, to approximate the value function $V_\theta(x,t)$ via self-supervised learning. 
The loss function for learning is composed of a PDE violation loss and a boundary condition loss with a trade-off parameter $\lambda$:
%
%
\begin{equation}
\label{eq: DeepReach_loss}
\begin{aligned}
& h\left(x_i, t_i ; \theta\right)=h_{pde}\left(x_i, t_i ; \theta\right)+\lambda h_{bc}\left(x_i, t_i ; \theta\right), \\
& h_{pde}\left(x_i, t_i ; \theta\right)=\| \min \left\{D_t V_\theta\left(x_i, t_i\right)+H\left(x_i, t_i\right),\right. \\
& \qquad \qquad \left.l\left(x_i\right)-V_\theta\left(x_i, t_i\right)\right\} \| ,\\
& h_{bc}\left(x_i, t_i ; \theta\right)=\left\|V_{\theta}\left(x_i, t_i \right)-l\left(x_i \right)\right\| \mathds{1}\left(t_i=T\right).
\end{aligned}
\end{equation}
Here, $h_{pde}$ corresponds to how consistent the value function propagation is with the HJB-VI, whereas $h_{bc}$ attempts to impose correct value function approximation at the terminal time. 
Even though DeepReach is shown to be effective for high-dimensional problems, the solution quality depends heavily on the imposition of $h_{bc}$ and $\lambda$, as we demonstrate later in this paper. 
Our goal is to reduce these dependencies in order to obtain a more accurate safety value function.

\section{Exact Imposition of Boundary Constraints}
We now propose \textit{ExactBC}, a novel variant of DeepReach, to overcome the aforementioned key challenges.
The main innovation of ExactBC is to ensure that the learned value function inherently satisfies the time boundary condition of (\ref{eq: HJI-VI}) for all states at time $t=T$. 
\vspace{-0.5em}
\subsection{Value Function Approximation}
We approximate the overall value function as:
\begin{equation} \label{eqn:bc_variant}
    V_\theta(x,t)=l(x)+ g_\theta(x,t), 
\end{equation}
where $l(x)$ is the target function encoding the failure set $\mathcal{L}$ and $g_\theta(x,t)$ can be any function parameterized by $\theta$ such that the $\forall x \in \mathbb{R}^n, \ g_\theta(x,T)=0$. The above structure guarantees that $\forall x \in \mathbb{R}^n, \ V_\theta(x,T)=l(x)$, thereby ensuring that the safety boundary constraints are exactly satisfied.
In this paper, we choose $g_\theta(x,t)$ as the product of the time factor, $(T-t)$, and the function $O_\theta(x,t)$ is given by a NN with parameters $\theta$. 
Thus, the overall value function can be represented as:
\begin{equation}
    V_\theta(x,t)=l(x)+ (T-t)* O_\theta(x,t).
\end{equation}
%
ExactBC has a few key advantages over DeepReach: first, ExactBC eliminates the need to learn the boundary condition explicitly; this is especially useful when the boundary value function is complex, making the learning process particularly challenging for DeepReach. 
Secondly, the NN now needs to minimize only a single loss function (corresponding to the HJB-VI), as the boundary condition is already exactly imposed, regardless of what the network learns. 
%
%
Specifically, the new loss function to learn the NN parameters is:
%
%
\begin{equation}\label{eq: loss}
\begin{aligned}
& h\left(x_i, t_i ; \theta\right)=h_{pde}\left(x_i, t_i ; \theta\right) \\
&=\| \min \left\{D_t V_\theta\left(x_i, t_i\right)+H\left(x_i, t_i\right), l\left(x_i\right)-V_\theta\left(x_i, t_i\right)\right\} \| \\
&=\| \min \left\{(T-t_i) D_t O_\theta\left(x_i, t_i\right) - O_\theta\left(x_i, t_i\right)+H\left(x_i, t_i\right),\right. \\
& \qquad \qquad \left.l\left(x_i\right)-V_\theta\left(x_i, t_i\right)\right\} \|, \\
& H\left(x_i, t_i\right)=\max_{u \in \mathcal{U}} \langle (T-t_i)\nabla O(x_i,t)+ \nabla l(x_i) ,f(x_i,u) \rangle.
\end{aligned}
\end{equation}
Compared to \eqref{eq: DeepReach_loss}, ExactBC eliminates the training uncertainty stemming from the choice of $\lambda$.
This also improves the convergence of stochastic gradient descent during training~\cite{chen2020comparison}. 
Finally, in ExactBC, the boundary value serves as a static and accurate ground-truth signal that anchors and stabilizes the self-supervised learning process. 
In contrast, in DeepReach, the learned boundary value dynamically varies during the training, causing instabilities in the value function.

\subsection{Training Scheme}\label{Prterain_Strat}
Since the solution of HJB-VI essentially propagates backward over time, i.e., the value function at time $t$ depends on the value function at time $t+\Delta t$. Similar to DeepReach, we adopt a curriculum training scheme over the time horizon to train the network. 
Specifically, the time interval for data sampling grows linearly from $[T,T]$ to $[0,T]$ during training. 
This curriculum training scheme allows the NN to learn accurate value function approximation starting from the terminal boundary conditions and slowly propagating it backward using the HJB-VI. 
However, as evident from ($\ref{eq: loss}$), the terminal time gradient of the value function depends on the parameter initialization of the DNN:
%
\begin{equation}
\label{eq: init_loss}
    D_t V_\theta\left(x_i, t\right)|_{t=T} = - O_\theta\left(x_i, T\right)
\end{equation}
This implies that poor weight initializations of the DNN can adversely impact the terminal time gradients, which may drive the learning process towards an undesirable local minimum. 
To address this issue, we introduce a pretraining strategy for the NN before performing the curriculum training using the following loss: 
\begin{equation}
\label{eq: pretrain_loss}
    h_{pretrain}\left(x_i,T ; \theta\right)=\left\|O_\theta\left(x_i, T\right)\right\|.
\end{equation}
%
%
Intuitively, the pretraining phase aims to simultaneously suppress the NN output and the time gradients at the terminal time. This smooths out the gradients and terminal boundary value function to mitigate the training instabilities that can stem from abnormally large initial values of the NN. In addition to this, it also makes the learned value function agnostic to the DNN's stochastic parameter initialization.

\begin{remark}
Note that both the pretraining and training phases utilize only a single loss function. The pretraining phase employs the pretraining loss detailed in \eqref{eq: pretrain_loss}, while the training phase utilizes the training loss derived in \eqref{eq: loss}.
\end{remark}

\section{Experiments} 
In this section, we conduct a comparative study on the performance of ExactBC on several reachability problems of varying complexity in terms of dynamics, state dimensionality, and safety boundary conditions.
\vspace{0.5em}

\noindent \textbf{\textit{Baselines and Training Details.}}
We compare ExactBC against DeepReach. For DeepReach, we further use the adaptive loss rebalancing scheme proposed in \cite{wang2021understanding} to reduce the impact of $\lambda$ on the learned value function.
In addition to DeepReach, we compare ExactBC against a variant of DeepReach that learns the residual between the value function and the boundary condition but does not impose the boundary conditions exactly: $V_\theta(x,t)=l(x)+ O_\theta(x,t)$. We call this variant DiffModel.
DiffModel reduces the learning burden on the NN as the network now only needs to learn the residual. 
Finally, as an ablation study, we also compare against a version of ExactBC that does not use the pertaining scheme described in Sec. \ref{Prterain_Strat}.

\begin{table*}[h!]
    \centering
    \vspace{1.0em}
    \resizebox{\textwidth}{!}{\begin{tabular}{c|c|c|c|c}
     \hline
      Method &\textbf{Rimless Wheel}   & \textbf{Bicycle Robot}  &\textbf{Rocket Landing}&  \textbf{Multi-Aircraft System} \\
    \hline
    DeepReach  & 44.1\% $\pm$ 4\%&  41.4\% $\pm$ 12.1\%&  0.4\% $\pm$ 0.1\%&  62.3\% $\pm$ 7.2\%  \\
      DiffModel &  41.1\% $\pm$ 1\%&  35.4\% $\pm$ 6.2\%&  11.7\% $\pm$ 7.1\%& 55.1\% $\pm$ 26.9\%\\
         ExactBC (Ours) &  \textbf{51.4\% $\pm$ 3\%}& \textbf{61.1\% $\pm$ 7.9\%}& \textbf{33.4\% $\pm$ 0.2\%} & \textbf{68.6\% $\pm$ 2.1\%}\\
         ExactBC without pretraining & 50.1\% $\pm$ 6\%&   58.1\% $\pm$ 7.1\%&  33.1\% $\pm$ 0.3\%& 68.3\% $\pm$ 6.3\%\\
     \hline 
    \end{tabular}}
    \caption{\small{Average recovered volume of the safe set ($\mu_{s}$) across the five seeds at a 99.99\% safety level. 
    The higher the $\mu_{s}$, the better the method.
    Best variant is marked in bold for each system. ExactBC consistently leads to a higher safe volume, especially when the underlying safety constraints are complex.}}
    \label{tab:BRT_Volume_Numbers}
\vspace{-0.5em}
\end{table*}

\begin{figure*}[h]
		\centering
         {\includegraphics[width=1.0\textwidth]{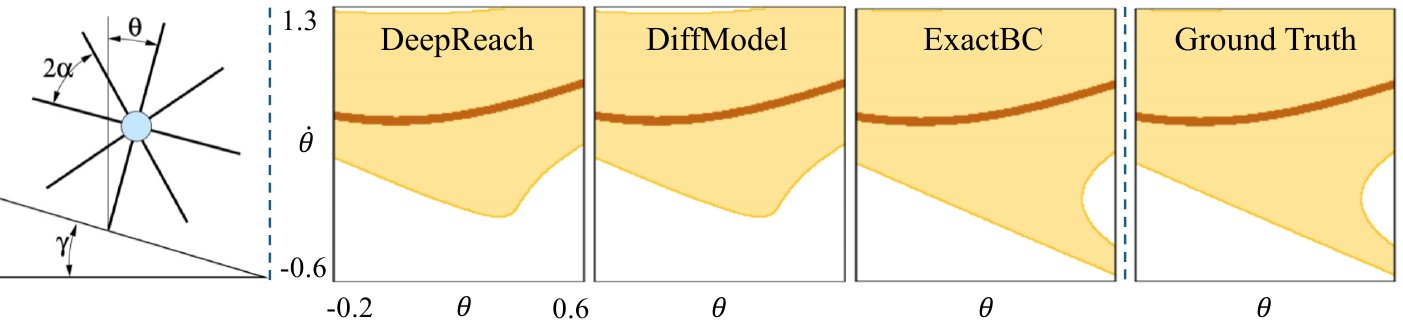}}
 	\caption{\small{(Left) Rimless wheel system. (Right) Trained and ground truth BRT slices. The brown and yellow areas represent the limit cycle (the target set) and the BRT respectively. ExactBC recovers a higher BRT volume than DeepReach and DiffModel, and its BRT is closely aligned with the ground truth BRT.}}
    \vspace{-1.8em}
 	\label{Plot_Rimless_BRT}
\end{figure*}

For all our case studies and variants, we utilize an NVIDIA GeForce RTX 3090 GPU for training and use consistent training settings: we use a 3-layer, fully connected, feedforward NN with a sinusoidal activation function and hidden layer size of 512. We perform curriculum training using the Adam optimizer with a learning rate of $2\times 10^{-5}$. 
%
In each training iteration, 65000 data points are uniformly sampled from the (normalized) state space (normalized to [-1,1]).
For each case study, we ran each variant with five different random seeds to capture the variance.
%
%
\vspace{0.5em}

\noindent \textbf{\textit{Evaluation Metric.}}
Our key evaluation metric is the volume of the safe set. For liveness problems, this corresponds to the volume of the BRT, whereas for safety problems, it represents the volume of the complement of the BRT. Generally, a larger volume of the safe set indicates better performance. To quantify this, we sample $N$ distinct states from the state space and count the number of safe states, i.e., the states that fall within the learned BRT for liveness problems or the number of states outside the learned BRT for safety problems, denoted as $n_{s}$. The percentage volume of the safe set is given as $\mu_{s} = 100 \times \frac{n_{s}}{N}$. We choose a large value of $N = 3 \times 10^6$ to attain samples from a significant portion of the state space.

Since the ground truth BRT is not available for some of our case studies (due to their high dimensionality), we use the verification method in \cite{pmlr-v242-lin24a} to compute the safe set volume. 
\cite{pmlr-v242-lin24a} provides a high-confidence bound $\delta$ on the learning error using conformal prediction. 
This results in a correction of the value function by the computed bound.
The corrected value function is then used to compute the BRT (or the complement of the BRT for safety problems).
We use a safety level of $99.99\%$ in all our case studies, resulting in a safety violation of almost 
$10^{-4}$ in the recovered safe set.
This recovered safe set is then used in the computation of $\mu_{s}$.
We refer interested readers to \cite{pmlr-v242-lin24a} for more details on the verification method.

\subsection{A Rimless Wheel System}\label{Rimless_Wheel}
%
         
%
\begin{figure*}[h!]
		\centering
         \includegraphics[width=1.0\textwidth]{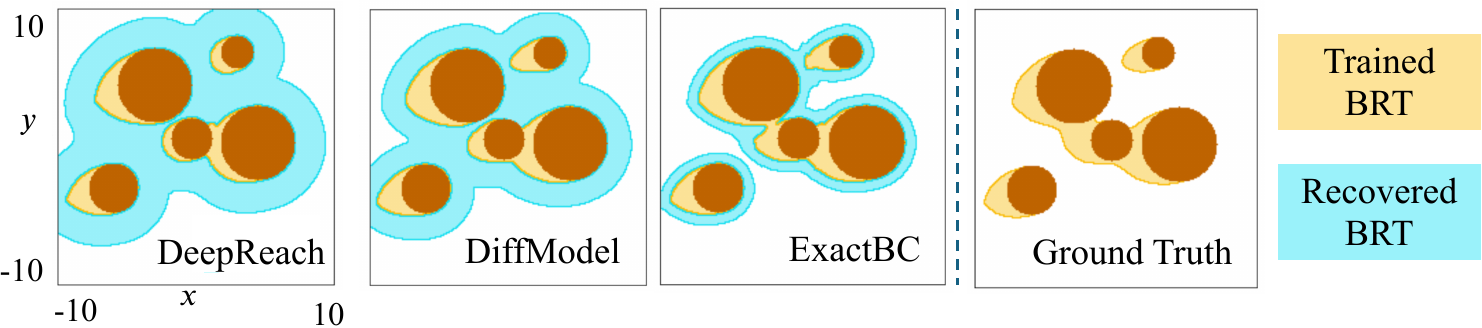}
 	\caption{\small{Ground truth, trained, and recovered BRT slices for the bicycle robot. 
  The recovered BRT is computed using the robust verification strategy at a safety level of 99.99\%.
  These slices correspond to $v=3.6 \text{m/s}$, $\theta=0.25 \text{rad}$ and $\psi=0$. Due to the complex boundary conditions in this case, ExactBC learns a more accurate value function and recovers a significantly higher BRT volume than DeepReach.}}
  \vspace{-1.8em}
 	\label{fig:Dubins_Recovery}
\end{figure*}
We first consider a 2D rimless wheel system \cite{choi2022computation}. 
The system comprises a central mass with multiple spokes extending radially outward and lacks control input. 
At any given moment, one spoke is pinned to the ground, causing the system to follow the dynamics of a simple pendulum, \( f(\theta, \dot{\theta}) =  [\dot{\theta}, \sin(\theta)]  \), where $\theta$ is the angle of contact of the spoke (see Fig. \ref{Plot_Rimless_BRT}). 
%
The rimless wheel model is often used as a simplified model of bipedal walking on an inclined plane.
Our goal is to reach and follow a stable periodic limit cycle (or gait) of the rimless wheel. Thus, the problem can also be thought of as computing the Region-of-Attraction (ROA) of the limit cycle.
In this case, the BRT or ROA represents the set of all initial configurations of the wheel from which it can stabilize to and follow the periodic gait.
From a reachability perspective, this problem can be formulated as a liveness problem, where the target set is given by:
\begin{equation*}
\mathcal{L} = \{x: \cos(\theta) + \frac{1}{2}\dot{\theta}^2 = E\},
\end{equation*}
where $E$ represents the total energy of the system. 

The complexity of this system arises from the fact that when an unpinned spoke makes contact with the ground, an inelastic collision takes place, causing an instantaneous change in \( \dot{\theta} \). Following the collision, the previously pinned spoke is released, and the newly contacted spoke becomes pinned. This causes an instantaneous switch in the system's states (also known as \textit{state resets}). The switching surface, $\mathcal{S}$, characterizes the set of states at which a state reset occurs, while the reset map, $\Delta$, specifies the mapping between the states before the reset and those after the reset. This can be mathematically expressed as:
\begin{equation*}
\mathcal{S} = \left\{x~|~\theta = \alpha + \gamma \right\}, \quad x^{-} = \Delta(x^{+}) = \begin{bmatrix}
2\gamma - \theta \\
\cos(2\alpha) \, \dot{\theta}
\end{bmatrix}
\end{equation*}
%
where $\gamma$ represents the incline of the surface on which the wheel rotates, and $2\alpha$ denotes the angle between two consecutive spokes. 
These state resets introduce stiffness in the system dynamics, making it harder to compute an accurate value function.

For our case study, we use $\gamma = 0.2$, $\alpha = 0.4$, and $E = 1.132$.
Given the low dimensionality of this system, we also compute the ground truth BRT using the Level Set Toolbox \cite{mitchell2004toolbox} that computes the value function over a 201 $\times$ 201 grid over [-0.2, 0.6] $\times$ [-1.3, 0.6] state space.
%
For learning-based approaches, we applied 50K pretraining iterations for all variants with a pretraining phase, followed by curriculum training on all the variants for 200K iterations. The BRT computation required 1.5 hours on an NVIDIA RTX 3090.

Fig.~\ref{Plot_Rimless_BRT} illustrates the obtained BRT slices for $T=6.3s$, and the BRT volume for different methods is compared in Table \ref{tab:BRT_Volume_Numbers}.
As evident from the figure, ExactBC obtains a much more accurate approximation of the BRT (it recovers $98.1\%$ of the ground truth volume, compared to $88.8\%$ and $83.92\%$ captured by DeepReach and DiffModel, respectively).
An interesting observation from these figures is that ExactBC accurately retrieves the BRT volume in regions near the switching boundary ($\theta=0.6$), whereas both DeepReach and DiffModel fail to do so. This can be explained by the fact that the dynamics are particularly stiff near the switching region, making the training process further susceptible to learning errors. 
Comparing the BRT volume in Table~\ref{tab:BRT_Volume_Numbers} indicates that 
ExactBC, when used without pretraining, demonstrates a relatively higher standard deviation compared to when it is pretrained. This can be attributed to poor initialization in some of the seeds, resulting in a relatively lower recovered volume. The proposed pretraining strategy reduces the variance and stabilizes the results across all seeds in such cases.

\subsection{Bicycle Robot in a Cluttered Environment}\label{Dubins5D}
For our next example, we consider a bicycle robot trying to avoid collisions in a cluttered environment containing five obstacles of different sizes (shown in Brown in Fig. \ref{fig:Dubins_Recovery}). 
The dynamics of the system are given as:
\begin{equation*}
\dot{p}_x = v \cos \theta,~\dot{p}_y = v \sin \theta,~\dot{v} = a,\\ 
~\dot{\theta} = \frac{v}{L} \tan \psi,~\dot{\psi} = \omega,
\end{equation*}
where $(p_x, p_y)$ is the position, $v$ is the forward speed, $\theta$ is the heading, $\psi$ is the steering angle, and $L = 1 \, \text{m}$ is the wheelbase length. The controls are acceleration, $a \in [-2, 2] \, \text{m/s}^2$ and steering angular velocity, $\omega \in [-2, 2] \, \text{rad/s}$. 
The failure set in this case is given by all the states that are in close vicinity of any of the obstacles:  
\begin{equation*}
\mathcal{L} = \{x: \min_{i \in \{1,2..5\}}(||(p_x, p_y) - (O_{x_i}, O_{y_i})|| - R_i) \leq 0\}, 
\end{equation*} 
where, $(O_{x_i}, O_{y_i})$ and $R_i$ are the center and radius of the $i$th obstacle, respectively. The time horizon, $T$, is $1s$.

%
We compute the BRT for this system using different learning-based variants, as well as the Level Set Toolbox. A grid size of $51 \times 51 \times 12 \times 12 \times 51$ was used for the Level Set Toolbox computation, yielding a coarse approximation of the ground truth BRT. Beyond that, the system runs out of RAM. This highlights the necessity of learning-based methods for efficient BRT computation in even moderate dimensional robotic systems. Each variant with a pretraining phase underwent 50K pretraining iterations, followed by 200K curriculum training iterations. The BRT computation required 2 hours on an NVIDIA RTX 3090.

Fig.~\ref{fig:Dubins_Recovery} shows the $xy$ slices of the ground truth BRT, the learned BRTs (yellow), as well as the recovered BRTs after using the verification method in \cite{pmlr-v242-lin24a} (cyan). 
As expected, because of the learning errors, the cyan region is always bigger than the yellow region, ultimately giving us the verified safe region in the white.
Additionally, as shown in Table~\ref{tab:BRT_Volume_Numbers}, ExactBC achieves an average recovered volume of $\mu_s = 61.12\%$. This performance is significantly superior compared to DeepReach and DiffModel, which recover volumes of $41.4\%$ and $35.4\%$, respectively. 
These results are interesting because when compared with the ground truth BRT, the learned ExactBC and DeepReach BRTs are comparable (ExactBC retrieves $95.05\%$ of the ground truth volume, whereas DeepReach retrieves $92.39\%$). 
Yet, their recovered volume after correction is significantly different ($61.12\%$ vs $41.4\%$).
These results may be attributed to the particularly complex boundary condition in this case, such as the presence of narrow gaps between obstacles, which makes it particularly challenging to obtain an accurate value function in this region, requiring a higher correction in the resultant value function for providing assurances (i.e., a bigger cyan region and a smaller white region in this case). 
%
These results highlight the benefit of using ExactBC for problems having complex boundary conditions.
%
\subsection{Autonomous Rocket Landing}\label{RLand}

\begin{figure*}[h!]
		\centering
         \includegraphics[width=1.0\textwidth]{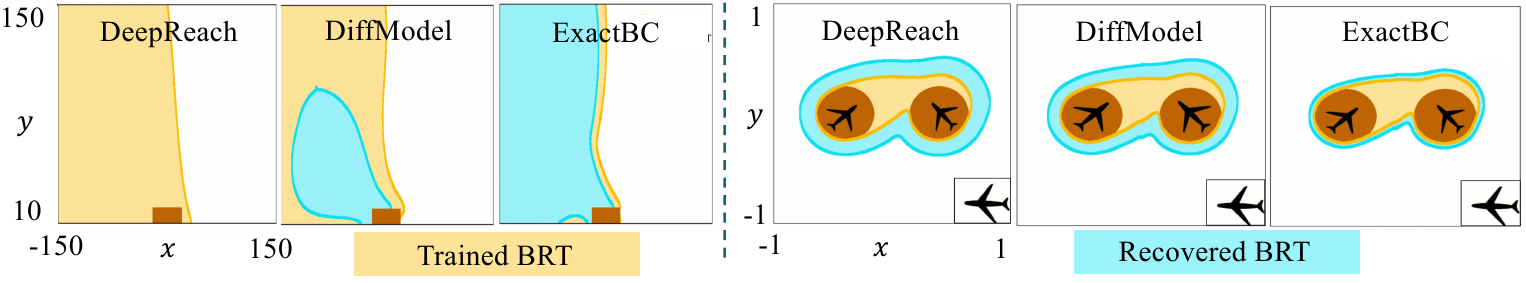}
 	\caption{\small{Trained and recovered BRT slices for (Left) rocket landing and (Right) multi-aircraft collision avoidance system. For the rocket landing system, DeepReach is not able to recover any volume that is verifiably safe, while DiffModel is only able to recover a small fraction of the volume. On the other hand, ExactBC is consistently able to recover a significant portion of the learned volume, which illustrates the increase in learning accuracy over the other baselines.}}
  \vspace{-1.8em}
 	\label{fig:RLand_Slices}
\end{figure*}
%
%
%
Next, we consider an autonomous rocket landing system, where a rocket must land safely on a rectangular pad within $T = 1s$. The system states include position $(x,y)$, heading $\theta$, velocity $(v_x,v_y)$, and angular velocity $\omega$. The rocket is controlled via torque inputs $\tau_1, \tau_2$, bounded within $[\tau_{min}, \tau_{max}]$.
The rocket dynamics are given as:
\begin{gather}
\qquad \qquad \dot{x} = v_x, \quad 
\dot{y} = v_y, \quad 
\dot{\theta} = \omega, \quad
\dot\omega = k \tau_1, \nonumber \\
\dot{v_x} = \tau_1 cos\theta - \tau_2 sin\theta, \nonumber \quad 
\dot{v_y} = \tau_1 sin\theta + \tau_2 cos\theta - g,
\end{gather}
where $g$ is gravitational acceleration and $k$ is a constant multiplier. The rectangular landing zone is of side length 20m centered around the origin (shown as the brown area in Figure~\ref{fig:RLand_Slices}). 
Hence, the target set is given as:
\begin{equation*}
\mathcal{L} = \{x, y: |x| \leq 20, y \leq 20\}.
\end{equation*}
%
For this example, we used 100K pertaining iterations for all the variants that have a pretraining phase followed by curriculum training on all the variants for 100K iterations. The BRT computation required 2.5 hours. This 6D system is beyond the computational capabilities of grid-based methods, so we don't compute a ground truth BRT in this case.
Once trained, the value functions were corrected using the method in \cite{pmlr-v242-lin24a} as described earlier.
%

Fig.~\ref{fig:RLand_Slices} shows the $xy$ slices of the learned BRT (yellow) as well as the slice of the recovered BRT after verification (cyan) for different variants. 
DeepReach completely fails on this problem and is not able to certify any part of the learned volume for any of the five seeds. This is because of high learning errors in certain parts of the state space, which leads to a very conservative error bound. 
The DiffModel shows only minor improvement over DeepReach as it is able to recover an average volume fraction of $11.7\%$, indicating that residual learning alone is not sufficient to obtain accurate value functions. 
Its recovered volume also has a high standard deviation of $7.1\%$ across the five seeds, making it very susceptible to the NN's initialization and, consequently, unreliable.
Specifically, in this case, the target set is very small; thus, a small error in estimating the boundary condition could lead to inaccurate BRT. Such scenarios highlight the utility of using ExactBC.
Indeed, ExactBC, with and without pretraining, can recover, on average, a much higher verified safe volume of $33.1\%$ and $33.4\%$, respectively. 

\subsection{Multi-Aircraft Collision Avoidance} \label{TV_avoidance}
%
%
We now consider a 9D three-aircraft collision avoidance problem.
%
%
%
%
For each aircraft, the states include its position $[x, y]$ and heading $\theta$. 
The control is a one-dimensional steering control $|u_i| \leq \bar{u}$. The velocity, $v$, of the aircrafts is equal and constant.
The dynamics of each aircraft are: 
\begin{equation*}
\begin{aligned}
&\quad \dot{x_i} = v cos\theta_{i}, \quad 
\dot{y_i} = vsin\theta_{i}, \quad 
\dot{\theta_i} = u_i \quad i=1,2,3,
\end{aligned}
\end{equation*}
which represent approximate aircraft dynamics at cruising altitude.
The failure set is given as the set of states where any two aircrafts are in close proximity of each other: 
\begin{equation*}
\mathcal{L} = \{x: \min\{d(C_1, C_2), d(C_2, C_3), d(C_1, C_3)\}\leq R\} 
\end{equation*}
where, $d(C_i, C_j)$ represents the distance between aircrafts $i, j$.
For our experiment, we use $v=0.6$, $\bar{u} = 1.1~rad/s$, and $R=0.25$. The time horizon, $T$, is $1s$.
We used 60K pertaining iterations for all variants with a pretraining phase followed by curriculum training on all the variants for 100K iterations. The BRT computation required 4 hours.
%
%

Fig.~\ref{fig:RLand_Slices} shows the trained and recovered BRT slices for different variants, and Table~\ref{tab:BRT_Volume_Numbers} presents the average recovered volume for each variant after verification across all seeds. 
The DeepReach solution yields an average recoverable volume of $62.3\%$. 
DiffModel's performance is rather poor with a high standard deviation of $26.9\%$, which can be attributed to its exceptionally poor performance for few seeds, where it failed to recover any volume.
This, again, validates the hypothesis that residual learning alone is not sufficient.
%
In contrast, ExactBC achieves a higher average recovered volume of $68.6\%$, outperforming both methods. ExactBC without pretraining attains a similar average volume but shows a much higher standard deviation, demonstrating the stabilization effects of the proposed pretraining scheme. 
 \label{experiments}
\vspace{-1.2em}
\section{Conclusion and Future Work}
In this work, we propose ExactBC, a novel variant of DeepReach that exactly imposes the safety boundary conditions during the learning process. ExactBC mitigates the learning inaccuracies introduced by imposing boundary conditions as soft constraints. Additionally, ExactBC learns the value function by optimizing over a single PDE loss term, thereby eliminating the need to weigh different losses correctly. Through case studies on challenging reachability problems, we demonstrate that ExactBC leads to significant improvement in the value function and BRT accuracy.

Despite its advantages, ExactBC is not without limitations. Notably, it struggles with non-differentiable boundary functions. This challenge arises because the reformulated PDE depends on the boundary function’s gradient, and non-differentiability can result in a noisy learning signal. Moreover, ExactBC does not offer computational benefits over Vanilla DeepReach and other variants, as they all require the same training time. Future works will focus on addressing these limitations by improving computational efficiency and the handling of non-differentiable boundary functions.   

\bibliographystyle{IEEEtran}
\bibliography{reference}

\begin{thebibliography}{10}
\providecommand{\url}[1]{#1}
\csname url@rmstyle\endcsname
\providecommand{\newblock}{\relax}
\providecommand{\bibinfo}[2]{#2}
\providecommand\BIBentrySTDinterwordspacing{\spaceskip=0pt\relax}
\providecommand\BIBentryALTinterwordstretchfactor{4}
\providecommand\BIBentryALTinterwordspacing{\spaceskip=\fontdimen2\font plus
\BIBentryALTinterwordstretchfactor\fontdimen3\font minus \fontdimen4\font\relax}
\providecommand\BIBforeignlanguage[2]{{%
\expandafter\ifx\csname l@#1\endcsname\relax
\typeout{** WARNING: IEEEtran.bst: No hyphenation pattern has been}%
\typeout{** loaded for the language `#1'. Using the pattern for}%
\typeout{** the default language instead.}%
\else
\language=\csname l@#1\endcsname
\fi
#2}}

\bibitem{8263977}
S.~Bansal, M.~Chen, S.~Herbert, and C.~J. Tomlin, ``Hamilton-jacobi reachability: A brief overview and recent advances,'' in \emph{IEEE 56th Annual Conference on Decision and Control (CDC)}, 2017, pp. 2242--2253.

\bibitem{mitchell2005time}
I.~M. Mitchell, A.~M. Bayen, and C.~J. Tomlin, ``A time-dependent hamilton-jacobi formulation of reachable sets for continuous dynamic games,'' \emph{IEEE Transactions on automatic control}, vol.~50, no.~7, pp. 947--957, 2005.

\bibitem{lygeros2004reachability}
J.~Lygeros, ``On reachability and minimum cost optimal control,'' \emph{Automatica}, vol.~40, no.~6, pp. 917--927, 2004.

\bibitem{fisac2015reach}
J.~F. Fisac, M.~Chen, C.~J. Tomlin, and S.~S. Sastry, ``Reach-avoid problems with time-varying dynamics, targets and constraints,'' in \emph{Proceedings of the 18th international conference on hybrid systems: computation and control}, 2015, pp. 11--20.

\bibitem{liao2024improved}
W.~Liao, T.~Liang, P.~Xiong, C.~Wang, A.~Song, and P.~X. Liu, ``An improved level set method for reachability problems in differential games,'' \emph{IEEE Transactions on Systems, Man, and Cybernetics: Systems}, 2024.

\bibitem{darbon2016algorithms}
J.~Darbon and S.~Osher, ``Algorithms for overcoming the curse of dimensionality for certain hamilton--jacobi equations arising in control theory and elsewhere,'' \emph{Research in the Mathematical Sciences}, vol.~3, no.~1, p.~19, 2016.

\bibitem{chow2017algorithm}
Y.~T. Chow, J.~Darbon, S.~Osher, and W.~Yin, ``Algorithm for overcoming the curse of dimensionality for time-dependent non-convex hamilton--jacobi equations arising from optimal control and differential games problems,'' \emph{Journal of Scientific Computing}, vol.~73, pp. 617--643, 2017.

\bibitem{lee2020hopf}
D.~Lee and C.~J. Tomlin, ``A hopf-lax formula in hamilton--jacobi analysis of reach-avoid problems,'' \emph{IEEE Control Systems Letters}, vol.~5, no.~3, pp. 1055--1060, 2020.

\bibitem{coogan2015efficient}
S.~Coogan and M.~Arcak, ``Efficient finite abstraction of mixed monotone systems,'' in \emph{Proceedings of the 18th International Conference on Hybrid Systems: Computation and Control}, 2015, pp. 58--67.

\bibitem{chen2015exact}
M.~Chen and C.~J. Tomlin, ``Exact and efficient hamilton-jacobi reachability for decoupled systems,'' in \emph{54th IEEE Conference on Decision and Control (CDC)}, 2015, pp. 1297--1303.

\bibitem{10365682}
C.~He, Z.~Gong, M.~Chen, and S.~Herbert, ``Efficient and guaranteed hamilton–jacobi reachability via self-contained subsystem decomposition and admissible control sets,'' \emph{IEEE Control Systems Letters}, vol.~7, pp. 3824--3829, 2023.

\bibitem{9561949}
S.~Bansal and C.~J. Tomlin, ``Deepreach: A deep learning approach to high-dimensional reachability,'' in \emph{2021 IEEE International Conference on Robotics and Automation (ICRA)}, pp. 1817--1824.

\bibitem{darbon2020overcoming}
J.~Darbon, G.~P. Langlois, and T.~Meng, ``Overcoming the curse of dimensionality for some hamilton--jacobi partial differential equations via neural network architectures,'' \emph{Research in the Mathematical Sciences}, vol.~7, pp. 1--50, 2020.

\bibitem{niarchos2006neural}
K.~Niarchos and J.~Lygeros, ``A neural approximation to continuous time reachability computations,'' in \emph{Proceedings of the 45th IEEE Conference on Decision and Control}, 2006, pp. 6313--6318.

\bibitem{djeridane2006neural}
B.~Djeridane and J.~Lygeros, ``Neural approximation of pde solutions: An application to reachability computations,'' in \emph{Proceedings of the 45th IEEE Conference on Decision and Control}, 2006, pp. 3034--3039.

\bibitem{rubiesroyo2017recursive}
V.~Rubies-Royo and C.~Tomlin, ``Recursive regression with neural networks: Approximating the hji pde solution,'' \emph{arXiv preprint arXiv:1611.02739}, 2016.

\bibitem{jiang2016using}
F.~Jiang, G.~Chou, M.~Chen, and C.~J. Tomlin, ``Using neural networks to compute approximate and guaranteed feasible hamilton-jacobi-bellman pde solutions,'' \emph{arXiv preprint arXiv:1611.03158}, 2016.

\bibitem{raissi2019physics}
M.~Raissi, P.~Perdikaris, and G.~E. Karniadakis, ``Physics-informed neural networks: A deep learning framework for solving forward and inverse problems involving nonlinear partial differential equations,'' \emph{Journal of Computational physics}, vol. 378, pp. 686--707, 2019.

\bibitem{saad2022guiding}
N.~Saad, G.~Gupta, S.~Alizadeh, and D.~C. Maddix, ``Guiding continuous operator learning through physics-based boundary constraints,'' \emph{arXiv preprint arXiv:2212.07477}, 2022.

\bibitem{li2020fourier}
Z.~Li, N.~Kovachki, K.~Azizzadenesheli, B.~Liu, K.~Bhattacharya, A.~Stuart, and A.~Anandkumar, ``Fourier neural operator for parametric partial differential equations,'' \emph{arXiv preprint arXiv:2010.08895}, 2020.

\bibitem{li2021physics}
Z.~Li, H.~Zheng, N.~Kovachki, D.~Jin, H.~Chen, B.~Liu, K.~Azizzadenesheli, and A.~Anandkumar, ``Physics-informed neural operator for learning partial differential equations,'' \emph{arXiv preprint arXiv:2111.03794}, 2021.

\bibitem{sukumar2022exact}
N.~Sukumar and A.~Srivastava, ``Exact imposition of boundary conditions with distance functions in physics-informed deep neural networks,'' \emph{Computer Methods in Applied Mechanics and Engineering}, vol. 389, p. 114333, 2022.

\bibitem{lagaris1998artificial}
I.~E. Lagaris, A.~Likas, and D.~I. Fotiadis, ``Artificial neural networks for solving ordinary and partial differential equations,'' \emph{IEEE transactions on neural networks}, vol.~9, no.~5, pp. 987--1000, 1998.

\bibitem{bokanowski2010reachability}
O.~Bokanowski, N.~Forcadel, and H.~Zidani, ``Reachability and minimal times for state constrained nonlinear problems without any controllability assumption,'' \emph{SIAM Journal on Control and Optimization}, vol.~48, no.~7, pp. 4292--4316, 2010.

\bibitem{mitchell2004toolbox}
I.~Mitchell, ``A toolbox of level set methods,'' \emph{http://www. cs. ubc. ca/mitchell/ToolboxLS/toolboxLS.pdf}, 2004.

\bibitem{pythonhjtoolbox}
E.~Schmerling, ``{hj\_reachability: Hamilton-Jacobi reachability analysis in JAX},'' \emph{https://github.com/StanfordASL/hj\_reachability}, 2021.

\bibitem{chen2020comparison}
J.~Chen, R.~Du, and K.~Wu, ``A comparison study of deep galerkin method and deep ritz method for elliptic problems with different boundary conditions,'' \emph{arXiv preprint arXiv:2005.04554}, 2020.

\bibitem{wang2021understanding}
S.~Wang, Y.~Teng, and P.~Perdikaris, ``Understanding and mitigating gradient flow pathologies in physics-informed neural networks,'' \emph{SIAM Journal on Scientific Computing}, vol.~43, no.~5, pp. A3055--A3081, 2021.

\bibitem{pmlr-v242-lin24a}
\BIBentryALTinterwordspacing
A.~Lin and S.~Bansal, ``Verification of neural reachable tubes via scenario optimization and conformal prediction,'' in \emph{Proceedings of the 6th Annual Learning for Dynamics \& Control Conference}, ser. Proceedings of Machine Learning Research, A.~Abate, M.~Cannon, K.~Margellos, and A.~Papachristodoulou, Eds., vol. 242.\hskip 1em plus 0.5em minus 0.4em\relax PMLR, 15--17 Jul 2024, pp. 719--731. [Online]. Available: \url{https://proceedings.mlr.press/v242/lin24a.html}
\BIBentrySTDinterwordspacing

\bibitem{choi2022computation}
J.~J. Choi, A.~Agrawal, K.~Sreenath, C.~J. Tomlin, and S.~Bansal, ``Computation of regions of attraction for hybrid limit cycles using reachability: An application to walking robots,'' \emph{IEEE Robotics and Automation Letters}, vol.~7, no.~2, pp. 4504--4511, 2022.

\end{thebibliography}
\balance

\end{document}